# Handling of uncertainty in medical data using machine learning and probability theory techniques: A review of 30 years (1991-2020)

Roohallah Alizadehsani[1,*], Mohamad Roshanzamir[2], Sadiq Hussain[3], Abbas Khosravi[1], Afsaneh Koohestani[1], Mohammad Hossein Zangooei[4], Moloud Abdar[1], Adham Beykikhoshk[5], Afshin Shoeibi[6,7], Assef Zare[8], Maryam Panahiazar[9], Saeid Nahavandi[1], Dipti Srinivasan[10], Amir F. Atiya[11], U. Rajendra Acharya[12,13,14]

[1]Institute for Intelligent Systems Research and Innovations (IISRI), Deakin University, Geelong, Australia. [2] Department of Engineering, Fasa Branch, Islamic Azad University, Post Box No 364, Fasa, Fars 7461789818, Iran. [3] System Administrator, Dibrugarh University, Assam 786004, India. [4]University of Texas at Dallas, USA. [5]Applied Artificial Intelligence Institute, Deakin University, Geelong, Australia. [6]Computer Engineering Department, Ferdowsi University of Mashhad, Mashhad, Iran. [7]Faculty of Electrical and Computer Engineering, Biomedical Data Acquisition Lab, K. N. Toosi University of Technology, Tehran, Iran. [8] Faculty of Electrical Engineering, Gonabad Branch, Islamic Azad University, Gonabad, Iran. [9]Institute for Computational Health Sciences, University of California, San Francisco, USA. [10]Dept. of Electrical and Computer Engineering, National University of Singapore, Singapore 117576. [11]Department of Computer Engineering, Faculty of Engineering, Cairo University, Cairo 12613, Egypt. [12]Department of Electronics and Computer Engineering, Ngee Ann Polytechnic, Singapore. [13] Department of Biomedical Engineering, School of Science and Technology, Singapore University of Social Sciences, Singapore, [14] Department of Bioinformatics and Medical Engineering, Asia University, Taiwan.

**Running Title:** Handling of Uncertainty in Medicine

[*] **Corresponding author:** Roohallah Alizadehsani, Institute for Intelligent Systems Research and Innovations (IISRI), Deakin University, Geelong, Australia.

E-mail: ralizadehsani@deakin.edu.au

**Conflict of interest statement**: The authors have no conflicts to disclose.

# Abstract

Understanding data and reaching valid conclusions are of paramount importance in the present era of big data. Machine learning and probability theory methods have widespread application for this purpose in different fields. One critically important yet less explored aspect is how data and model uncertainties are captured and analyzed. Proper quantification of uncertainty provides valuable information for optimal decision making. This paper reviewed related studies conducted in the last 30 years (from 1991 to 2020) in handling uncertainties in medical data using probability theory and machine learning techniques. Medical data is more prone to uncertainty due to the presence of noise in the data. So, it is very important to have clean medical data without any noise to get accurate diagnosis. The sources of noise in the medical data need to be known to address this issue. Based on the medical data obtained by the physician, diagnosis of disease, and treatment plan are prescribed. Hence, the uncertainty is growing in healthcare and there is limited knowledge to address these problems. We have little knowledge about the optimal treatment methods as there are many sources of uncertainty in medical science. Our findings indicate that there are few challenges to be addressed in handling the uncertainty in medical raw data and new models. In this work, we have summarized various methods employed to overcome this problem. Nowadays, application of novel deep learning techniques to deal such uncertainties have significantly increased.

## 1. Introduction

Machine learning is widely used in academia and industry to analyse big and complex datasets to uncover the hidden patterns and reach conclusive insights. It is already known that the performance of machine learning models has a close relationship not only with the selected algorithms but also depends on the nature of data. For example, having a significant amount of missing values and noise in the data can affect the results. Indeed, such uncertain data can be found in various fields such as energy systems (Gallagher et al. 2018; Soroudi and Amraee 2013), web (Nguyen et al. 2019), image (Shadman Roodposhti et al. 2019), disease and healthcare (Alizadehsani et al. 2019a; Alizadehsani et al. 2020; Alizadehsani et al. 2019b; Alizadehsani et al. 2019c; Arabasadi et al. 2017; Reamaroon et al. 2019), and the Internet of Things (Dilli et al. 2018). Hence, it can be argued that having such uncertainty in medical data makes the decision-making process difficult (Figure 1).

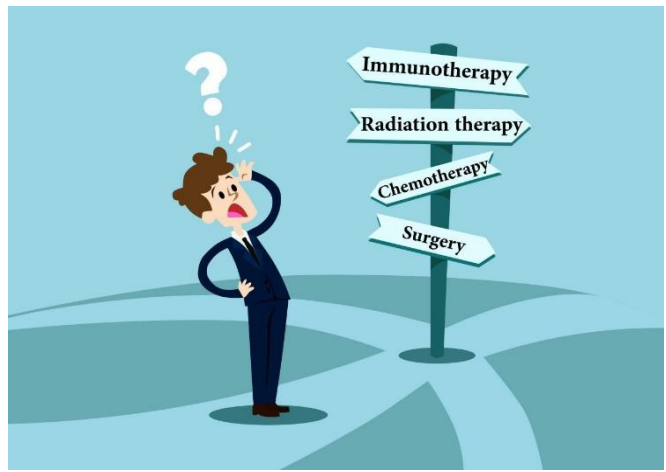

*Figure 1. Having uncertainty makes the decision-making process difficult*

Machine learning algorithms that can model uncertainty assist to reveal beneficial information for a better decision-making process. Generally, uncertainty may be due to two reasons: data (noise) uncertainty and model uncertainty (also called *epistemic uncertainty*) (Gal 2016). It is likely to have noise among labels due to measurement imprecision which may lead to *aleatoric uncertainty*. Meanwhile, model uncertainty can be divided into *two* main types: structure uncertainty and uncertainty in model parameters (Gal 2016). In structural uncertainty, we find out what model structure should be used and how we can specify our proposed model for either extrapolating and/or interpolating. In uncertainty in model parameters, the best model parameters will be chosen to accurately predict in a given dataset.  More specifically, Gal (Gal 2016) mentioned the importance of the model uncertainty in various fields such as biology, physics, and manufacturing.

For the first time, Renée Fox conducted few studies which showed the uncertainties faced by physicians during their training (Fox 1957; Fox 1980). After that, other researchers too acknowledged the central problematic nature of this problem. Although in health care, the importance of uncertainty has been growing, we have limited knowledge about how to solve many of these problems. The aim of this paper is to review the related studies in the domain of uncertainty quantification in the field of medical science. The results obtained showed that uncertainty is a common challenge among different raw data and various models. Moreover, most of the applied algorithms are Bayesian inference, fuzzy systems, Monte Carlo simulation, rough classification, Dempster–Shafer theory, and imprecise probability.

The organization of the other sections is described as follows. The search criteria used for finding papers is explained in Section 2. Section 3 provides more details about machine learning and probability theory methods used for handling uncertainty. Discussion and conclusion sections are presented in sections 4 and 5, respectively.

## 2. Search Criteria

To perform this review, we performed Google scholar search for the papers published between 1 January 1991 and 31 May 2020. We have obtained most of the published papers by IEEE, Elsevier and Springer. The search keywords used for this study is Bayesian inference OR fuzzy systems OR Monte Carlo simulation OR rough classification OR Dempster–Shafer theory OR Imprecise probability AND Medical Science.

Then, about 324 papers in English language are selected and those which do not make significant impacts were removed from the list. Hence 91 high quality papers indexed in Scopus or PubMed or having a large number of citations are considered. Then, we reviewed the references of all the selected papers to find more relevant papers. Finally, 74 papers are selected in this step to be added to 91 previously selected papers. Overall, 165 papers are investigated in this review. This procedure is illustrated in Figure 2. In Figure 3, the research trend in the past 30 years is shown. It is clear from the diagram that handling uncertainty in the medical data has received more attention in recent years.

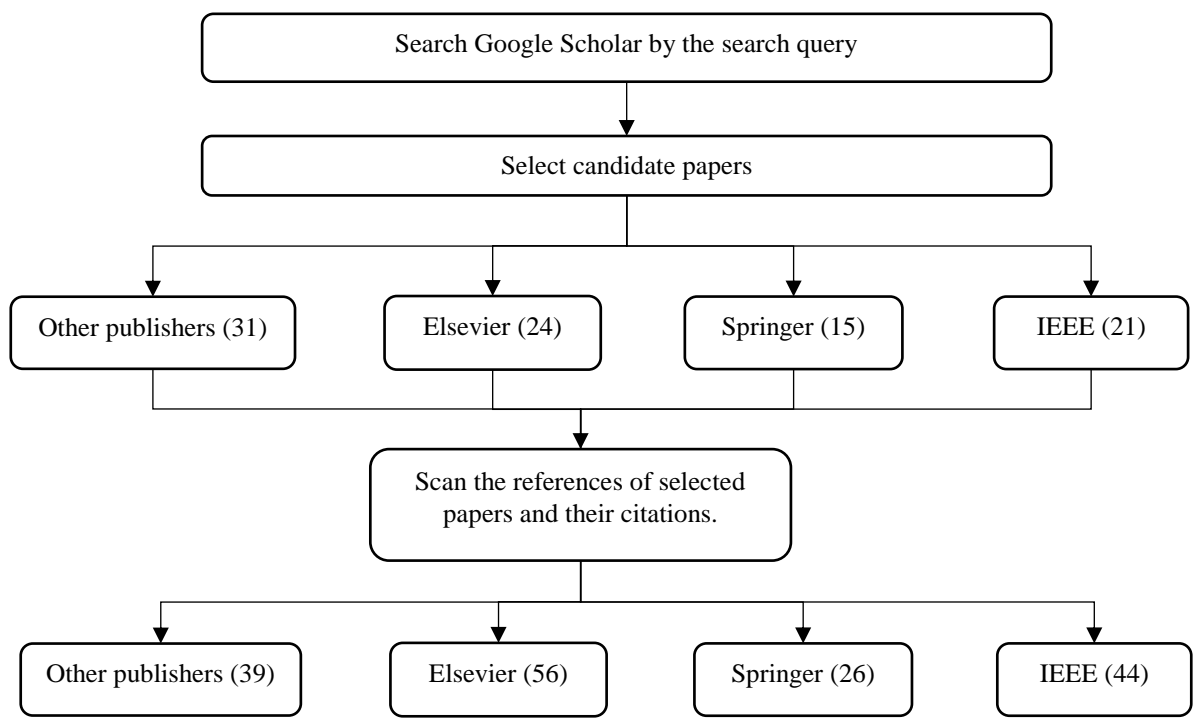

*Figure 2. Paper selection mechanism*

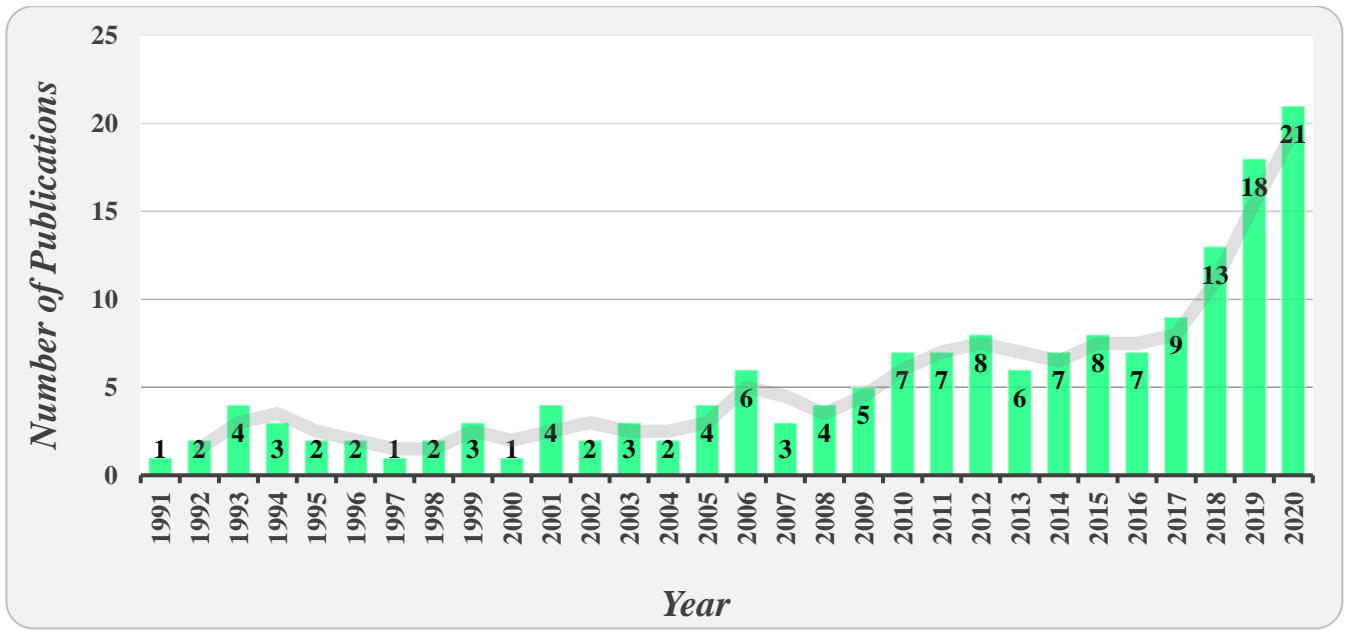

*Figure 3. Number of papers published in handling uncentainity in medical data between 1991 and 2020.*

## 3. Uncertainty Handling Algorithms in Medical Science

As it is shown in Figure 4, the most common algorithms in this field are Bayesian inference, fuzzy systems, Monte Carlo simulation, rough classification, Dempster–Shafer theory, and imprecise probability.

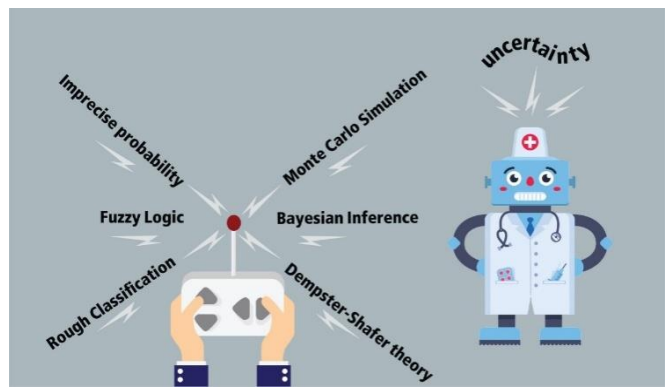

*Figure 4. The most common algorithms of uncertainty handling*

### 3.1. Bayesian inference (BI)

BI is the famous statistical inference techniques which utilizes Bayes' theorem in its inference mechanism (Akkoyun et al. 2020; Corani et al. 2013; Howle et al. 2017; Kourou et al. 2020; Seixas et al. 2014; Wang et al. 2019; Watabe et al. 2014). If there is more information or evidence, the probability of a hypothesis can be updated by utilizing Bayes' theorem (Ocampo et al. 2011).

Different probabilities are described by modelling experience or knowledge base using BI which is a sort of expert system. A fundamental model is indicated by these probabilities. Based on the fundamental model, a conclusion is acquired by repeating the Bayes' theorem using the inference engine. The following equation denotes the Bayes' theorem, using the events N and Y.

$$P(N/Y) = \frac{P(Y/N) * P(N)}{P(Y/N) * P(N) + P(Y/no\ N) * P(no\ N)} \quad (1)$$

where Y is particular combinations of signs and symptoms of a patient and N represents the fact "a patient suffers from a particular disease". By going through symptoms of Y, the probability of a person suffering from an ailment is equal to the probability of Y multiplied by the probability of the occurrence of the symptoms Y, whether the ailment prevails or not and it is shown as:

$$P(N/Y) = \frac{P(N \cap Y)}{P(Y)} \quad (2)$$

and

$$P(Y/N) = \frac{P(N \cap N)}{P(N)} \quad (3)$$

Therefore,
$$P(Y \cap N) = P(Y/N) * P(N) = P(N/Y) * P(Y) \qquad (4)$$

From which we get:
$$P(N/Y) = \frac{P(Y/N) * P(N)}{P(Y)} \qquad (5)$$

Bayes' theorem repeatedly is applied to find a likely diagnosis of the ailment and details are described structurally as above.

A priori probability of hypothesis, given it is false or true can be utilized to calculate the probability of specific hypothesis in this specific case. Hence in the case of element of evidence, Bayes' theorem can be represented based on indicative data and ailments as follows:

$$P(G/F) = \frac{PT * PJ}{PT * PJ + PM * (1 - PJ)} \qquad (6)$$

P(G) depicts the likelihood of occurring an ailment and it starts with P(G)=PJ for each ailment. P(G/F) is computed by acquiring the data from the user. The prior rule is applied when a symptomatic datum prevails. Otherwise, PT and PM are substituted by (1-PT) and (1-PM) by applying the same rule. A priori probability P(G) is substituted by P(G/F) as the outcome of each datum.

### 3.1.1. Related work based on bayesian inference

The accuracy of screening tests that were applied to identify antibodies to the human immunodeficiency virus with the help of Bayesian methods devised by Johnson et al. (Johnson and Gastwirth 1991). They tried to assess the incidence of the disease from the collected samples. Approximate predictive distributions for the number of future individuals that would test true positive was estimated by utilizing a novel sample or population of interest.

Robertson et al. (Robertson and DeHart 2010) proposed an accessible and agile adoption of Bayesian inference to design an expert system for medical diagnosis. It can be used by mid-level and low health workers in rural and remote locations. They suggested that the success of the expert system depended on the clinical interface for use in specific regions, rapid adaptation of the database, and by varying user skills.

Mazur (Mazur 2012) assessed the Bayesian inference in the context of medical decision making that had three key developments: (1) need for data recognition, (2) progress of inverse probability, and (3) development of probability. The author examined Bayesian inference development from the beginning with the effusive evidence of the clinician's sign to the work of Laplace, Jakob Bernoulli, and others.

In (Ashby 2006), Ashby examined the usage and applicability of Bayesian thinking in medical research. He reviewed the Bayesian statistics in medicine launched in 1982 to technologies associated with medical decision making such as molecular genetics, survival modeling, longitudinal modeling, spatial modeling, and evidence synthesis.

Suchard et al. (Suchard and Redelings 2006) proposed Bayesian posterior sampler called BAli-Phy that exploited Markov chain Monte Carlo to study phylogeny and the joint space of alignment given molecular sequence data. Their model automatically used information in shared deletions/insertions to help deduce phylogenies by utilizing more sophisticated substitution models in the alignment process.

The authors of (Mendoza-Blanco et al. 1996) utilized the theories of simulation-based techniques and missing-data analysis to develop a framework of Bayesian analysis to estimate the prevalence of immunodeficiency virus (HIV). Different practical considerations that arose from HIV screening was taken into account by their flexible techniques.

In (Huang et al. 2011), Huang et al. presented a Bayesian approach that mutually models three components. There were covariate, response, and time-to-event procedures linked thorough the random effects that demonstrated the vital individual-specific longitudinal procedures. The occurrence of CD4 covariate procedure having calculated errors in the HIV dynamic response was discussed and scrutinized disease progression by decreasing CD4/CD8 ratio to evaluate antiretroviral treatment.

The distributed and big data in the medical informatics platform framework of the Human Brain Project was analyzed by Melie-Garcia et al. (Melie-Garcia et al. 2018). They applied multiple linear regression (MLR) techniques. The Bayesian formalism that offered the armamentarium necessary to execute MLR techniques for distributed Big Data was employed by them. Their technique combined multimodal heterogeneous data coming from various hospitals and subjects around the world and recommended urbane ways that were extendable to other statistical models.

Galesic et al. (Galesic et al. 2009) investigated whether natural or conditional probabilities frequencies assisted precise measures in older persons and whether the results depended on numeracy skills vis-à-vis medical procedures.

The authors of (Johnston et al. 2015) devised novel generalizable theoretical and physically motivated model mitochondrial DNA (mtDNA) populations by assigning the first statistical comparison of proposed bottleneck mechanisms.

In (Huang et al. 2010), a hierarchical Bayesian modeling technique was presented to devise a method induced by an AIDS clinical study. Long-term virologic responses were characterized by integrating fully time-dependent drug efficacy, baseline covariates, pharmacokinetics, and drug resistance into the model. The experimental results showed that modeling virologic responses and HIV dynamics with consideration of baseline characteristics as well as time-varying clinical factors might be critical for HIV studies.

Henriquez et al. (Henriquez and Korpi-Steiner 2016) examined the Bayesian inference dilemma in medical analytics and asserted that there was a requirement for probabilistic reasoning tools which would be user-friendly in nature. These papers are briefly described in Table 1.

*Table 1. Brief description of papers which used Bayesian inference to handle uncertainty.*

| Authors | Year | Disease/Area | Work description |
|---|---|---|---|
| Lin et al. (Lin et al. 2020) | 2020 | Human dietary risk | Human dietary risk assessment using Bayesian inference |
| Zhou et al. (Zhou et al. 2020) | 2020 | Medical image reconstruction | Uncertainty quantification and Bayesian inference for the reconstruction of the medical image using Poisson data |
| Akkoyun et al. (Akkoyun et al. 2020) | 2020 | Abdominal aortic aneurysm | Abdominal aortic aneurysm prediction using two-step Bayesian inference |
| Magnusson et al. (Magnusson et al. 2019) | 2019 | Medical data analysis | Principal stratum estimand to examine the effect of treatment in a subgroup using Bayesian Inference |
| Lipkova et al. (Lipková et al. 2019) | 2019 | Brain Tumor | Radiotherapy blueprint for Glioblastoma by integrating Bayesian Inference, Multimodal Scans, and Tumor models |
| Flugge et al. (Flügge et al. 2019) | 2019 | Medical data analysis | Diagnostic inference and Knowledge representation utilizing Bayesian networks |
| Wang et al.(Wang et al. 2019) | 2019 | Lung Cancer | Lung Cancer patients' medical expenditure prediction utilizing Bayesian networks |
| Schultz et al. (Schultz et al. 2019) | 2019 | Medical Imaging | Uncertainty quantification in deformable image registration using Bayesian inference |
| Melie-Garcia et al. (Melie-Garcia et al. 2018) | 2018 | Neuro-imaging | Bayesian inference for Big and distributed Data in the Medical Analytics using multiple linear regression |
| Henriquez et al. (Henriquez and Korpi-Steiner 2016) | 2016 | Medical decision making | Bayesian inference dilemma in medical decision making |
| Johnston et al. (Johnston et al. 2015) | 2015 | Bioinformatics | MtDNA bottleneck mechanism using Bayesian inference and Stochastic modeling |
| Huang et al. (Huang et al. 2011) | 2011 | HIV | HIV dynamics with longitudinal data |
| Huang et al. (Huang et al. 2010) | 2010 | HIV | HIV dynamic differential equation models using hierarchical Bayesian inference |
| Robertson et al. (Robertson and DeHart 2010) | 2010 | Medical decision making | An accessible and agile adaptation of Bayesian inference to medical diagnostics for interior health workers |
| Galesic et al. (Galesic et al. 2009) | 2009 | Medical decision making | Medical screening tests' evaluation with natural frequencies of older people with low numeracy |
| Suchard et al. (Suchard and Redelings 2006) | 2006 | Bioinformatics | BAli-Phy: simultaneous Bayesian inference of alignment and phylogeny. |
| Mendoza-Blanco et al. (Mendoza-Blanco et al. 1996) | 1996 | HIV | A missing-data approach with simulation-based techniques on Bayesian inference prevalence related to HIV screening |
| Johnson et al. (Johnson and Gastwirth 1991) | 1991 | Medical decision making | Bayesian inference for medical screening tests |

*3.1.2. Related works based on Bayesian Deep Learning*

Bayesian neural networks (BNNs) are used to add uncertainty handling in models. As it is shown in Figure 5, in BNNs, it learns the parameters of a random variable instead of deterministic weights. Then, backpropagation is used to learn the parameters of the random variables.

In some researches, Bayesian neural network was used to handle uncertainty. Some of these types of researches will be explained in the following.

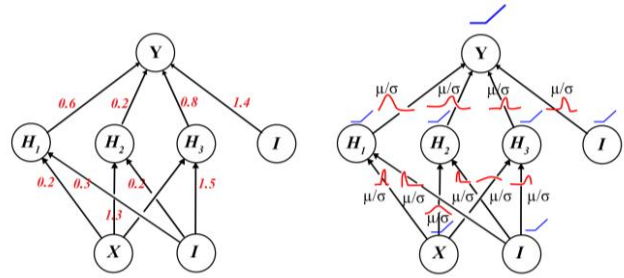

*Figure 5. Classical backpropagation sets fixed values as weights (left) while in Bayesian neural network a distribution is assigned to each weight (right).*

Kendall et al. (Kendall and Cipolla 2016) designed a visual relocalization system which has six degrees of freedom. It was a robust and real-time monocular system which is able to work indoors and outdoors scenes. It used the Bayesian convolution neural network (CNN) to obtain an accurate localization system on an outdoor dataset. Meanwhile, it was able to detect the presence of the scene in the input images. Kendall et al. (Kendall et al. 2015) proposed a learning algorithm based on deep learning for probabilistic semantic segmentation. The main goal of the algorithm was to understand the visual scene. It also can also take care of uncertainty during decision making. It was done by Monte Carlo sampling with dropout in the test phase to generate the posterior probability of the pixel labels. By modelling uncertainty, the segmentation performance improved by about 3%. Especially in the small dataset, the performance improved significantly because modelling uncertainty is more effective.

Although perception tasks such as object detection need human intelligence, a subsequent task that needs inference and reasoning needs more human intelligence. Nowadays, new algorithms like deep learning could extremely improve perception tasks (Ghassemi et al. 2020; Khodatars et al. 2020; Sharifrazi et al. 2020; Shoeibi et al. 2021; Shoeibi et al. 2020a; Shoeibi et al. 2020b). However, for higher-level inference, probabilistic graphical models are more powerful than other algorithms. Probabilistic graphical models are based on Bayesian reasoning. So, it seems that integrating deep learning and Bayesian models could handle both perception and inference problems which is called Bayesian deep learning. These two sections of the unified framework could boost each other. The object detection using deep learning improved the performance of a higher-level inference system. Then, the inference process feedback can enhance the object detection task. In (Wang and Yeung 2016), a general framework was proposed for Bayesian deep learning. Then, it was used in recommender or control systems.

In (Gal et al. 2017), an active learning framework was combined with Bayesian deep learning. Using the advantage of Bayesian convolutional neural network, they have achieved significant improvement in the existing active learning approaches, especially on high dimensional data. They tested their proposed algorithm on the MNIST dataset and also for skin cancer diagnosis. Two main branches for uncertainty modelling are aleatoric and epistemic. The former handles uncertainty in data and the later handles it in the model. Kendall et al. (Kendall and Gal 2017) focused on epistemic uncertainty in computer vision using Bayesian deep learning models. They combined aleatoric uncertainty with epistemic uncertainty. The framework was modelled with semantic segmentation and depth regression tasks. In addition, the formulated uncertainty resulted in a new loss function. This new loss function was robust to data uncertainty. In (Chen et al. 2013), a model was created by a hierarchical convolutional factor-analysis. The parameters of the layer-dependent model were calculated by variational Bayesian analysis and Gibbs sampler. An online edition of variational Bayesian was used to handle large-scale and streaming data. Using beta-Bernoulli distribution, they estimated the number of basic functions at each layer. This system was applied to image processing applications.

Instead of Gaussian process (GP) to model distributions over functions, Snoek et al. (Snoek et al. 2015) used neural networks and showed that the performance of their method could overtake state-of-the-art GP-based approaches. However, unlike GP which scales cubically, this method scales linearly with the number of data. Using this modification, they have achieved intractable degree of parallelism, rapidly finding other models using convolutional networks, and some other applications such as image caption generation. Nowadays, one of the states-of-the-art fields is object detection using a deep convolutional neural network (CNN). Using this method, Zhang et al. (Zhang et al. 2015) extracted discriminative features for categorization. They used for localization using a Bayesian optimization to detect the object bounding box. In (Gal and Ghahramani 2015), as labelled data is hard to be prepared, the authors applied the CNN on small data. However, in this case, CNN overfitted quickly. To overcome this problem, a type of Bayesian CNN was proposed that offered better robustness when there are small data. This was done by pacing the probability distribution on CNN kernels. Bernoulli variational distributions were used to approximate the created model. This way, they did not need additional model parameters. Theoretically, they cast dropout network training as approximate inference in the neural network. So, they used existing tools in deep learning without increasing the time complexity. Hence, it achieved considerable improvement in accuracy as compared to other prevailing techniques.

Bayesian parameter estimation for the deep neural networks is suboptimal for issues with small datasets. This problem also exists where accurate posterior predictive densities are needed. Monte Carlo method can be utilized as one of the solutions for this problem. However, this method needs to keep many copies for various parameters. It also need to make a prediction in many versions of the model. These two problems consume memory and time respectively. In the proposed method in (Korattikara et al. 2015), a more compact form of Monte Carlo approximation to the posterior predictive density is suggested. Then it was compared with two prevailing approaches: an approach depended on expectation propagation and variational Bayes. Authors claim that, their method performed better than both of these approaches.

In (Dahl et al. 2013), using a proposed method with ReLUs, the overall system's performance was enhanced by 4.2% with respect to deep neural networks (DNN) trained with sigmoid units and 14.4% with respect to the well-built Gaussian mixture / hidden Markov technique. Meanwhile, this method needs minimum hyper-parameter tuning using a regular Bayesian optimization code.

In (Louizos et al. 2017), the Bayesian method was used to handle this problem. Two novelties were proposed: the first one is that they pruned nodes instead of individual weights. Another novelty is that they used the posterior uncertainty to resolve the best possible fixed-point precision to decide the weights. These two factors could improve the performance of the system. One of the flexible probabilistic models is a Bayesian neural network with latent variables. These models can be used for the estimation of the network weights. Using these models, Depeweget al. (Depeweg et al. 2018) showed how to perform a decomposition of uncertainty for decision-making purposes. These methods can identify informative points of functions when these points had heteroscedastic and bimodal noise. By decomposition, they also define a risk-sensitive criterion for reinforcement learning. Using this learning method, a policy that balanced the expected cost, model-bias, and noise aversion can be found.

## 3.2. Fuzzy Logic

Mathematicians define a fuzzy set as a class of objects whose elements have a degree of membership which is determined by a membership function. Fuzzy logic (Zadeh 1988) is a method that its concepts are defined based on fuzzy set. In fuzzy logic, descriptive expressions are used for facts and rules expressions. The fuzzification operator is used to convert crisp values into fuzzy membership functions (Zadeh 1988). Defuzzifying operations are used to map fuzzy membership functions into standard numbers which are used for decision and control purpose. To convert a standard system to fuzzy system, the following three steps are done (Zadeh 1988).

1. Convert the inputs into fuzzy membership functions (Fuzzification).
2. Apply fuzzy rules to fuzzy input values to estimate the fuzzy outputs.
3. Convert the fuzzy outputs to standard outputs (Defuzzification).

Fuzzy logic maybe used to model uncertainty (Dervishi 2017; Khodabakhshi and Moradi 2017; Majeed Alneamy et al. 2019; Ornelas-Vences et al. 2017; Rundo et al. 2020; Sengur 2008; Toğaçar et al. 2020). However, there is a more powerful extension of it dubbed adaptive neuro-fuzzy inference system (ANFIS) that integrated the learning capability of fuzzy logic with neural networks to model uncertainty in expressiveness. Fuzzy logic is utilized to model uncertain scenarios and that model is learned by neural network. In the next section, it will be explained in detail.

### 3.2.1. ANFIS

The building block of ANFIS is Takagi–Sugeno fuzzy inference system (Karaboga and Kaya 2019). It is capable of capturing the benefit of fuzzy logic and artificial neural network both as it integrates two of them into a single framework. The IF-THEN fuzzy rules are utilized to learn approximate nonlinear functions in this inference system. The best parameters extracted from the genetic algorithm may be used to apply ANFIS in an optimal and efficient way. It is termed as a universal estimator and applies a situational aware intelligent energy management system. The architecture is comprised of five layers. The fuzzification and rule layers are the first and second layers respectively. The fourth layer takes the input from the third layer that normalized the values. The defuzzificated values are passed to the last layer that returns the final output.

The rule base comprises of a fuzzy if-then rules of Takagi and Sugeno's type as depicted below:

If x is A and y is B then z is f(x,y)

where z=f(x,y) is a crisp function in the consequent, and A and B are the fuzzy sets in the antecedents. A first order Sugeno fuzzy model is formed if f(x,y) is considered as first order polynomial. To form a first order two rule Sugeno fuzzy inference system, two rules may termed as below:

Rule 1: If x is $A_1$ and y is $B_1$ then $f_1 = p_1 x + q_1 y + r_1$

Rule 2: If x is $A_2$ and y is $B_2$ then $f_2 = p_2 x + q_2 y + r_2$

The type-3 fuzzy inference system as proposed by Takagi and Sugeno is used here. Its structure is shown in Figure 6 (Liu et al. 2019).

In this inference system, the input variables added by a constant term linearly form the output of each rule.

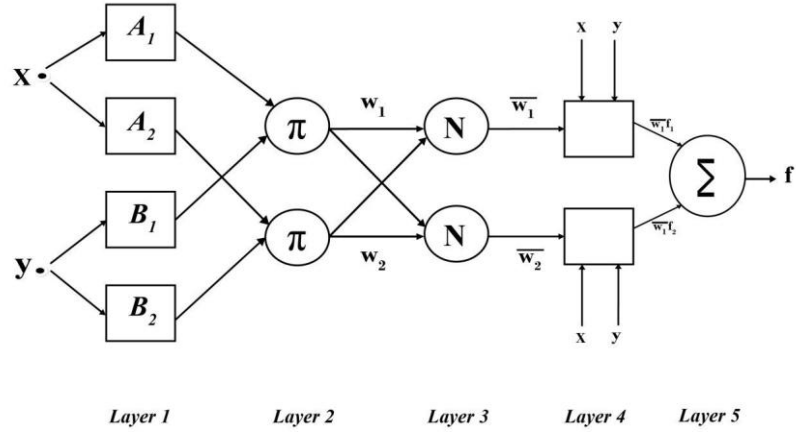

Figure 6. Type-3 ANFIS structure.

The individual layers of ANFIS structure are described below:

**Layer 1:** All the node i in layer 1 are adaptive with a node function

$$O_i^1 = \mu_{A_i}(x) \tag{7}$$

where, x is the input to node i, $\mu_{A_i}$ is the membership function of $A_i$ and $A_i$ is the linguistic variable related with this node function. The membership function is selected as below:

$$\mu_{A_i}(x) = \frac{1}{1 + \left[\left(\frac{x-c_i}{a_i}\right)^2\right]^{b_i}} \tag{8}$$

or

$$\mu_{A_i}(x) = \exp\left\{-\left(\frac{x-c_i}{a_i}\right)^2\right\} \tag{9}$$

where $\{a_i, b_i, c_i\}$ is the principle parameter set and x is the input.

**Layer 2:** The firing strength $\omega_i$ of a rule is measured by a fixed node that belongs to this layer. The product of each incoming signals to it is the output of this node and is depicted as below:

$$O_i^2 = \omega_i = \mu_{A_i}(x) \times \mu_{B_i}(y), i = 1,2 \tag{10}$$

**Layer 3:** Layer 3 is represented as below:

$$O_i^3 = \overline{\omega_i} = \frac{\omega_i}{\omega_1 + \omega_2}, i = 1,2 \tag{11}$$

**Layer 4:** Every node in this layer is adaptive layer with node function described as below:

$$O_i^4 = \overline{\omega_i} f_i = \overline{\omega_i}(p_i x + q_i y + r_i), i = 1,2 \tag{12}$$

where $\overline{\omega_i}$ is the output of the layer 3 and $\{p_i, q_i, r_i\}$ is the consequent parameter set.

Layer 5: It is consists of only one fixed node that measures the overall output as the summation of all incoming signals. It is described as below:

$$O_i^5 = \sum_i \overline{\omega_i} f_i = \frac{\sum_i \omega_i f_i}{\sum_i \omega_i} \tag{13}$$

As shown in figure 6, the first and fourth layers are adaptive layers. Three adjustable parameters {$a_i$, $b_i$, $c_i$} exist and associated with the input membership functions in the first layer. These parameters are dubbed as premise parameters. Three adjustable parameters {$p_i$, $q_i$, $r_i$} exist in the fourth layer. These are termed as consequent parameters.

### 3.2.2. Related Works based on ANFIS

The authors of (Turabieh et al. 2019) proposed a Dynamic ANFIS (D-ANFIS) to handle the missing values in the application used for the internet of medical things. This way they can overcome the potential problems that may occur. In (Ziasabounchi and Askerzade 2014), authors devised an ANFIS-based classifier to detect the degree of heart disease based on characteristic data of patients. The prediction model utilized seven variables as input. Their empirical results yielded an accuracy of 92.3% with k-fold cross-validation strategy.

Early diagnosis of chronic kidney disease (CKD) can prevent or reduce the progression of renal failure. Yadollahpour et al. (Yadollahpour et al. 2018) proposed an ANFIS-based expert medical decision support system (MDSS) to predict the timeframe of renal damage. They considered the glomerular filtration rate (GFR) as the biological marker of renal failure. The ANFIS model utilized current GFR and diabetes mellitus as underlying disease, diastolic blood pressure, and weight as the effective factors in renal failure prediction. Their model predicted accurately the GFR variations using long future timeframes.

The authors of (Salah et al. 2013) designed a helping device for elderly people. The device is called E-JUST assistive device (EJAD). Inertial sensors and a motion capture system recognized human posture. The EJAD comprises of an active walker and a robot arm. Fuzzy system applying ANFIS was trained by the modified IMUs to the right posture of the patient.

One of the autoimmune ailments is rheumatoid arthritis (RA) that directs to significant mortality and morbidity. Özkan et al. (Özkan et al. 2010) extracted the key features from the left and right hand Ulnar artery Doppler (UAD) signals for the detection of rheumatoid arthritis disease by using multiple signal classification (MUSIC) techniques. ANFIS used the features derived from left and right hand UAD signals for the classification of RA. The hybrid model comprised of ANFIS and MUSIC techniques demonstrated accuracies of 91.25% using left hand UAD signals and 95% using the right hand UAD in the early recognition of rheumatoid arthritis disease.

In (Yang et al. 2014), the authors devised a risk assessment prediction model for coronary heart disease by optimizing linear discriminant analysis (LDA) and ANFIS methods. They applied a Korean Survey dataset. Their technique yielded a lofty prediction rate of 80.2 % in preventing coronary heart disease.

The authors of (Polat and Güneş 2006) tried to use k-nearest neighbor (k-NN) and principal component analysis (PCA) based weighted for pre-processing of data, and ANFIS to diagnose thyroid disease. As the first step, they tried to reduce the dimension of thyroid disease data from five features to only two features using PCA. Then based on k-NN, a pre-processing phase was applied on data and finally, ANFIS was used for diagnosing thyroid disease.

In (Kumar et al. 2003), a recursive method was used for the fuzzy system online learning employing Tikhonov regularization. This system was based on the recursive solution of a nonlinear least-squares problem. In (Ghazavi and Liao 2008), a collection of fuzzy systems was used. A fuzzy k-NN algorithm, fuzzy clustering, and ANFIS were applied on medical datasets. The results showed that feature selection is an important phase for time reduction and accuracy increasing of the proposed algorithm.

Papageorgiou (Papageorgiou 2011) proposed the Fuzzy inference map to handle the problem of risk analysis and assessment of pulmonary infections in the hospital. Their proposed method was a soft computing algorithm that could deal with situations such as uncertain descriptions. In (Nguyen et al. 2015), a new method which was a combination of wavelet transformation (WT) and interval type-2 fuzzy logic system (IT2FLS) was used for medical noisy and high-dimensional data classification. IT2FLS could handle uncertainty and noise in complex medical data. Their method could serve as a decision support system in clinical settings. A brief description of these works are summarized in Table 2.

*Table 2. Brief description of papers which used ANFIS to handle uncertainty.*

| Authors | Year | Disease/Area | Work description |
|---|---|---|---|
| Das et al. (Das et al. 2020) | 2020 | Medical disease analysis | Feature Extraction Model using Neuro-Fuzzy for classification |
| Tiwari et al. (Tiwari et al. 2020) | 2020 | Lung Cancer | Fuzzy Inference System for detection of lung cancer |
| Kour et al. (Kour et al. 2020) | 2020 | Medical disease analysis | Neuro-fuzzy systems for prediction and classification of different types of diseases |
| Vidhya et al. (Vidhya and Shanmugalakshmi 2020) | 2020 | Medical disease analysis | Modified-ANFIS using various disease analysis based on medical Big Data |
| Kaur et al. (Ranjit et al. 2020) | 2020 | Knee Diseases | Knee Diseases Prediction using adaptive and improved ANFIS |
| Hekmat et al. (Hekmat et al. 2020) | 2020 | Acute kidney Injury | Risk factors, prevalence, and early outcome analysis of acute kidney injury |
| Sood et al. (Sood et al. 2020) | 2020 | dengue fever | LDA-ANFIS based dengue fever risk assessment framework |
| Sujatha et al. (Sujatha et al. 2020) | 2020 | Breast cancer | Micro calcifications in breast identification utilizing ANFIS |
| Liu et al. (Liu et al. 2019) | 2019 | Prostate Cancer | Using a fuzzy inference system, prostate cancer was predicted |
| Turabieh et al. (Turabieh et al. 2019) | 2019 | Breast cancer | A D-ANFIS is used to handle the missing values in the application used for the Internet of Medical Things |
| Mori et al. (Mori et al.) | 2019 | Medical decision making | Extracting the relationship between input and output of the learning data using fuzzy rules. |
| Medeiros et al. (de Medeiros et al. 2017) | 2017 | Medical decision making | Real-time medical diagnosis using a fuzzy inference system. |
| Nguyen et al. (Nguyen et al. 2015) | 2015 | Breast cancer | A new classifier based on the type-2 fuzzy logic system for breast cancer diagnosis |
| Azar et al. (Azar and Hassanien 2015) | 2014 | Breast cancer | Medical big data dimensionality reduction using a neuro-fuzzy classifier |
| Papageorgiou et al. (Papageorgiou 2011) | 2011 | Pulmonary infections | A fuzzy approach to handle uncertainty for decision making and modeling medical knowledge |
| Ghazavi et al. (Ghazavi and Liao 2008) | 2008 | Diabetes | Fuzzy modeling methods used for diabetes diagnosis |
| Polat et al. (Polat and Güneş 2006) | 2006 | Thyroid disease | k-NN and ANFIS based hybrid medical decision support system |
| Kumar et al. (Kumar et al. 2003) | 2003 | Physiological data | A new method using Sugeno type fuzzy inference system applied to physiological data |

## *3.3. Monte Carlo simulation*

It is a technique that is applied to compute the uncertainty in variety of fields such as financial problems and project management. Overall, this technique is used for predicting models and estimate the probability of a system outcome. It is used when we are faced with a system having random variables. Monte Carlo (MC) simulation helps to predict all potential results of a system. This way, the users can take better decisions according to the risk and uncertainty of the system. This is why this method is referred to as a probability simulation by some researchers (Arnaud Doucet et al. 2001). The main idea is that one can solve the problem using randomness. This method is commonly used in physical and mathematical problems when other methods are hard to be used. They are usually used for optimization and numerical integration (Grzymala-Busse 1988).

### 3.3.1. Monte Carlo method

MC techniques are a subset of computational algorithms that utilize the procedure of repeated random sampling to obtain numerical estimations of unknown parameters. They assess the impact of risk and allow the modelling of critical situations where many random variables are engaged (Ghobadi et al. 2020; Hecquet et al. 2007; Jeeva and Singh 2015; Precharattana et al. 2011). The uses of the method are exceptionally widespread and have led numerous innovative discoveries in the fields of finance, game theory, and physics (Liesenfeld and Richard 2001). A broad range of Monte Carlo methods shares the generality that they depend on random number generation to crack deterministic problems (Koistinen 2010).

**Basic Principle Monte Carlo Integration**

Let us say $f$ is a density that we simulate from and are interested in the expectation (Liesenfeld and Richard 2001)

$$I = \int h(x)f(x)dx = Eh(X) \tag{14}$$

Suppose from set $Y_i=h(X_i)$ and density f, we simulate $X_1$, $X_2$, …and then the sequence $Y_1$, $Y_2$, … is i.i.d. and $EY_i=Eh(X_i)= \int h(x)f(x)dx = I$. We compute the N values $h(X_1),…, h(X_N)$ to acquire the estimate

$$\hat{I}_N = \frac{1}{N}\sum_{i=1}^{N} h(X_i) \tag{15}$$

By the SLLN, as N increases, $\hat{I}_N$ converges to I with the condition $E|h(x)| < \infty$. We are free to select N as large as available computer time in Monte Carlo simulations. It is easy to select the standard deviation and variance of the estimator. If the variance of the average is denoted as below then the variance of the single term h(X) is finite (Liesenfeld and Richard 2001).

$$var\ \hat{I}_N = \frac{1}{N} var h(X) \tag{16}$$

This is termed as Monte Carlo variance, simulation variance, or sampling variance of the estimator $\hat{I}_N$. The accuracy of the $\hat{I}_N$ can be measured more meaningfully by applying the square root of the variance. The square root of the variance of an estimator is also dubbed as standard error. The standard error of a Monte Carlo estimate is termed as Monte Carlo the standard error, simulation standard error, or sampling standard error. The Monte Carlo standard error is of the order $1/\sqrt{N}$, since

$$\sqrt{var\ \hat{I}_N} = \frac{1}{\sqrt{N}}\sqrt{var\ h(X)} \tag{17}$$

The population variance or theoretical variance var h(X) that is required in both equations 16 and 17 is generally unknown. It can be however measured by the sample variance of $h(X_i)$ values,

$$s^2 = \widehat{var}\ h(X) = \frac{1}{N-1}\sum_{i=1}^{N}(h(X_i) - \hat{I}_N)^2 \tag{18}$$

We achieve an approximate 100(1-α) % confidence interval for I namely

$$\hat{I}_N \pm z_{1-\frac{\alpha}{2}}\frac{s}{\sqrt{N}} \tag{19}$$

### 3.3.2. RELATED WORKS BASED ON MONTE CARLO SIMULATION (MCS)

In (Papadimitroulas et al. 2012), extensive validation of MCS toolkit named as GATE is used to estimate dose point kernels (DPKs). It is widely used for many medical physics applications such as patient dosimetry and Computed Tomography (CT) image simulation. The results are compared with reference data to produce a total DPKs complete dataset for radionuclides in nuclear medicine.

In (Downes et al. 2009), the X-ray volume imager was modelled utilizing a novel Monte Carlo (MC) code. In this work, a novel constituent module was devised to precisely mould the unit's bowtie filter. The results showed good agreement between measurement and MC.

In (Chen et al. 2009), a new method is proposed to investigate scatter in CT breast imaging. It was done by comparing the distribution of measured scatter to those simulated using the Monte Carlo simulation toolkit (named as Gate). The results of scatter measurements were compared to previous ones. It was observed that from a non-breast source, a significant scatter can arise. The validated Monte Carlo simulation toolkit was also used to describe the scatter in different X-ray settings as well as for various breast sizes.

Authors in (Bush et al. 2008) generated an automated system named VIMC-Arc based on MC. Their designed system requires minimal user input like patient ID, requested dose uncertainty, and required voxel size. This system can be a great platform for the analysis of dosimetric problems with varying degrees of tissue inhomogeneity.

Jia at al. (Jia et al. 2011) developed a MC dose calculation package that used graphics processing unit (GPU) capabilities. They named their package as gDPM v2.0. They could achieve high computational power using GPU architecture without decreasing the accuracy. They tested their system on both phantoms and realistic patients using central processing unit (CPU) and GPU simulations. There was no significant difference between the accuracy of these simulations. However, GPU simulations were much faster than CPUs.

A good review on the capability of GATE Monte Carlo simulation was done in (Sarrut et al. 2014) for dosimetry applications and radiation therapy. The GATE MC simulation platform works based on the GEANT4 toolkit. Many applications that used GATE for radiotherapy simulations were reviewed in this research. An important feature of the GATE which makes it easy to model both treatment and image acquisitions within the same system was also emphasized.

In (Lee et al. 2012), a database was introduced which was established by a complete organ-effective dose i.e. (33 organs and tissues) based on CT scanner MC simulation. The test cases ranged from new born to 15-year-old male and female. The achieved results were compared with three existing researches. This comparison showed that phantoms using realistic anatomy is essential for better accuracy in CT organ dosimetry.

A simple approach was suggested in (Wang and Leszczynski 2007) to evaluate the size and the shape of a linac's focal spot by comparing the profiles with the MC calculated ones from the profile of measured dose data. A brief description of these papers are summarized in Table 3.

*Table 3. Brief description of papers which used Monte Carlo simulation to handle uncertainty.*

| Authors | Year | Disease/Area | Work description |
| --- | --- | --- | --- |
| Tsai et al. (Tsai et al. 2020) | 2020 | DNA damages due to ionizing radiation | GPU-oriented microscopic Monte Carlo simulation tool |
| Aubin et al. (Aubin et al. 2020) | 2020 | Health occupation Education | Item Analysis and Examinee Cohort Size for Health occupation Education |
| Lee et al. (Lee 2020) | 2020 | Breast Cancer | Study of dual-head Compton camera with CZT/Si material for detection of Breast Cancer |
| Shih et al. (Shih et al. 2020) | 2020 | Dose assessment of a blood irradiator | Dose assessment of a blood irradiator utilizing MAGAT gel dosimeter and Monte Carlo simulation |
| Gasparini et al. (Gasparini et al. 2020) | 2020 | Medical Data Analysis | Monte Carlo simulation study for health care longitudinal data in Mixed-effects models |
| Sun et al. (Sun et al. 2020) | 2020 | Cancer | Monte Carlo simulation for verification of proton beam range |
| Lee et al. (Lee et al. 2012) | 2012 | Medical decision making | Monte Carlo simulation for medical decision making using an organ dose database |
| Jia et al. (Jia et al. 2011) | 2011 | Cancer | Radiotherapy dose calculation by using Monte Carlo simulation based on GPU |
| Downes et al. (Downes et al. 2009) | 2009 | Medical decision making | Calculating dose point kernels using the GATE Monte Carlo simulator and comparing it with reference data. Proposing a complete dataset of dose point kernels |
| Chen et al. (Chen et al. 2009) | 2009 | Breast cancer | Characterizing scatter in cone-beam CT breast imaging. |
| Bush et al. (Bush et al. 2008) | 2008 | Medical decision making | Rapid Arc radiotherapy delivery Monte Carlo simulation |
| Wang at al. (Wang and Leszczynski 2007) | 2007 | Medical decision making | Using MC simulation for focal spot size and shape estimation |
| Hérault et al. (Hérault et al. 2005) | 2005 | Ocular melanoma | Proton therapy platform MC simulation for ocular melanoma |

*3.4. Rough set theory (RST)*

After fuzzy, probability and evidence theories, a novel mathematical tool called RST [63] to deal with uncertain and inconsistent knowledge was proposed. The applications based on RST have increased in recent years as more researchers are attracted to this area (Chen 2013; Stokić et al. 2010; Wang et al. 2010; Zhang et al. 2010). In the artificial intelligence domain, it is one of the hot topics. It was originated from information model. The basic concept comprised of two stages. The first stage forms rules and concepts via classification of relational databases. The second stage is to mine knowledge through classification for approximation of the target and classification of the equivalence relation.

### 3.4.1. Formal Definition

Related data reasoning or analysis of algorithms and approximation of sets are critical research problems in the rough sets domain. Some basic concepts are described in this section. Let us consider an information system $I$ represented as the 4-tuple (Zhang et al. 2016)

$$I = <U, X, V, f>, X = C \cup D, \tag{20}$$

where subsets D and C are called decision attribute set and condition attribute set, X is a finite nonempty set of attributes, U is a finite non empty set of attributes. $V = \bigcup_{a \in X} V_a$, where $V_a$ is the value of attribute $a$, f: X>V is a description function and card $(V_a) > 1$.

**Definition 1.** (Indiscernible relation). An indiscernible relation ind(B) on the universe U given a subset of attribute set $B \subseteq X$ is defined as

$$ind(B) = \{(x, y) | (x, y) \in U^2, \forall_{b \in B} (b(x) = b(y))\} \tag{21}$$

The pair (U,$[x]_{ind(B)}$) is termed as approximation space, equivalence class of an object x is defined by $[x]_{ind(B)}$ or [x] if no confusion arises (Zhang et al. 2016).

**Definition 2.** (Upper and lower approximation sets). Let $I = <U, X, V, f>$ be an information set for a subset $Z \subseteq U$, its upper and lower approximation sets are termed respectively by (Zhang et al. 2016)

$$\underline{apr}(Z) = \{x \in U | [x] \cap Z \neq \emptyset\} \tag{22}$$

$$\overline{apr}(Z) = \{x \in U | [x] \in Z\} \tag{23}$$

Where [x] defines the equivalence class of x.

**Definition 3.** (Definable sets) Let $I = <U, X, V, f>$ be an information set for a target subset $Z \subseteq U$, and attribute subset $B \subseteq X$, if and only if $\overline{apr}(Z) = \underline{apr}(Z)$, Z is termed as definable set with respect to B.

**Definition 4.** (Rough Sets) Let $I = <U, X, V, f>$ be an information set for a target subset $Z \subseteq U$, and attribute subset $B \subseteq X$, if and only if $\overline{apr}(Z) \neq \underline{apr}(Z)$, Z is termed as rough set with respect to B.

**Definition 5.** (Roughness of rough sets) Let $I = <U, X, V, f>$ be an information set for a target subset $Z \subseteq U$, and attribute subset $B \subseteq X$, the roughness of set Z is denoted as below with respect to B,

$$P_B(Z) = 1 - |\underline{apr}(Z)|/|\overline{apr}(Z)| \tag{24}$$

Where |.| defines the cardinality of a finite set and $Z \neq \emptyset$.

Three disjoint regions of a rough set are illustrated in Figure 7 (Zhang et al. 2016). The boundary region leads to uncertainty in a rough set. The larger the boundary region will lead to the higher degree of uncertainty.

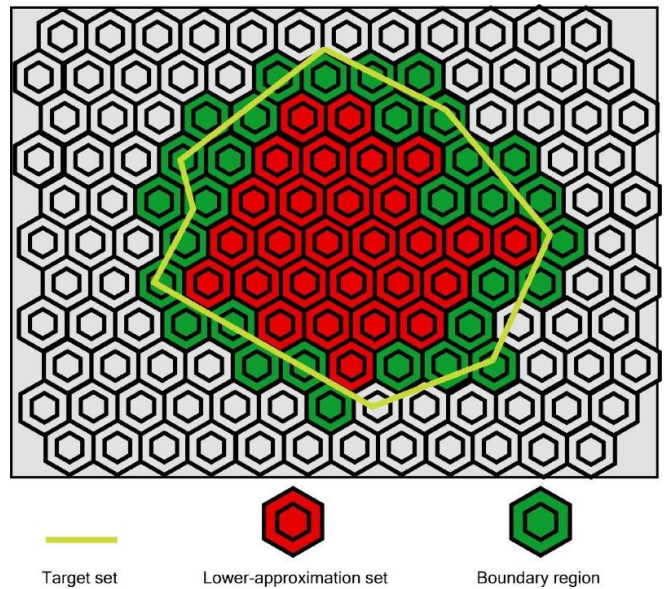

*Figure 7. Three disjoint regions of a rough set.*

### 3.4.2. Related Works based on RST

In (Kai-Quan 2002), Shi discussed S-rough set and their application in diagnosis of diseases. The author also presented the characteristics and structures of S-rough set.

Jiang et al. (Jiang et al. 2017) proposed a new model for sub-health diagnosis with reference to traditional Chinese medicine (TCM). Fuzzy weighted matrixes were generated and sub-health classification was done after extracting key features by utilizing fuzzy mathematics and rough sets. The novel method proved its efficacy when compared to other models in this domain.

In (Ningler et al. 2009), a novel approach by slightly modifying the variable precision rough set model was presented. In this research, Ningler et al. tested both the models using the dataset containing electroencephalogram of anesthetized and awake patients. Their technique produced smaller rule sets and achieved better cutback of features for inconsistent or noisy records despite of higher computational effort.

Wang et al. (Wang et al. 2010) devised a novel tumor classification technique based on neighborhood rough set and an ensemble of probabilistic neural network model based gene reduction. Gene ranking was applied to select informative genes and minimum gene subsets were chosen by reducing gene. Their method recorded competitive performance and was not sensitive to initially selected genes.

In (Tsumoto 1998a), authors proposed a rule induction technique which extracted classification rules as well as medical knowledge for the diagnosis of disease. The model was tested on three clinical datasets, whose results induced diagnostic rules correctly and estimated statistical measures as well.

Chou et al. (Chou et al. 2007) utilized rough set theory (RST) and self-organizing map (SOM) methods to evaluate laboratory test and drug utilization to extract knowledge from the raw data of cardiovascular disease patients. The model achieved an accuracy of 98% and detected the trend of patient's condition individually.

In (Tsumoto 1998b), the characteristics of expert rules were examined and a new method to derive plausible rules by utilizing *three* procedures was proposed. The proposed technique was evaluated on medical datasets and experts' decision processes were represented by induced rules. A brief description of these papers are summarized in Table 4.

*Table 4. Brief description of papers which used Rough classification to handle uncertainty.*

| Authors | Year | Disease/Area | Work description |
|---|---|---|---|
| Jain et al. (Jain and Kulkarni 2020) | 2020 | Computer-aided diagnosis | Multi-reduct classifier based on Rough set using medical data |
| Santra et al. (Santra et al. 2020) | 2020 | Low back pain | Low back pain management using Rough set based lattice structure |
| Bania et al. (Bania and Halder 2020) | 2020 | Medical data classification | R-Ensembler: Ensemble attribute selection using greedy rough sets |

| Li et al. (Li et al. 2020) | 2020 | Disaster classification | Medical rescue approaches analysis |
| --- | --- | --- | --- |
| Ahmed et al. (P and Acharjya 2019) | 2019 | Heart Disease | Heart Disease Diagnosis using Cuckoo search and Rough Set |
| Jiang et al. (Jiang et al. 2017) | 2017 | Sub-Health | Sub-health diagnosis based on rough set and fuzzy mathematics |
| Wang et al. (Wang et al. 2010) | 2010 | Tumor | Tumor classification |
| Ningler et al. (Ningler et al. 2009) | 2009 | Cardiovascular Disease | Adapted variable precision rough set techniques for EEG records |
| Chou et al. (Chou et al. 2007) | 2007 | Cardiovascular Disease | Drug utilization knowledge using rough set theory and self-organizing map |
| Shi (Kai-Quan 2002) | 2002 | Typhoid Fever | S-rough sets in diagnosis of disease |
| Tsumoto (Tsumoto 1998a) | 1998 | Headache and facial pain | Automated medical expert system rules' extraction |
| Tsumoto (Tsumoto 1998b) | 1998 | Muscle contraction headache | Experts' decision rules mining |

## *3.5. Dempster–Shafer theory (DST)*

P. Dempster and his student Glenn Shafer introduced Dempster Shafer Theory (Denźux 2016). The theory tried to overcome the limitations of Bayesian methods. Bayesian probability cannot describe ignorance and Bayesian theory concerns about single evidences. It is an evidence theory and it integrates all possible outcomes of the problem. The uncertainty in this model is as follows:
1. All possible outcomes are considered.
2. Belief will direct to believe in some likelihood by carrying out some evidences.
3. Plausibility will make evidence compatibility with possible outcomes.

### *3.5.1. Formal Definition*

All possible states of a system taken into account by the set termed as Z, the universe.

The set of all subsets of Z including empty set $\emptyset$ is represented by the power set $2^Z$.

Suppose, if Z={a,b}, then

$$2^Z = \{\emptyset, \{a\}, \{b\}, Z\} \tag{25}$$

Propositions are represented by the elements of the power set concerning the actual state of the system, by having all and only the state in which the proposition is true. A belief mass to each element is assigned by the theory of evidence. Formally, a function m: $2^Z \rightarrow [0,1]$ is termed as basic belief assignment, having two properties. First property is "the mass of the empty set is zero" i.e. $M(\emptyset) = 0$.

Secondly, the masses of the remaining members of the power set added up to total of 1.

$$\sum_{L \in 2^Z} m(L) = 1 \tag{26}$$

The upper and lower bounds of a probability interval can be described from the mass assignments. The interval comprised of the precise probability of a set and is bounded by two non-additive continuous measures called plausibility and belief.

$$bel(L) \leq P(L) \leq pl(L) \tag{27}$$

The belief bel (L) for a set L is denoted as the addition of all the masses of subsets of the set of interest.

$$bel(L) = \sum_{N \mid N \subseteq L} m(N) \tag{28}$$

The plausibility pl (L) is the addition of all the masses of the sets N that intersect the set of interest L.

$$pl(L) = \sum_{N \mid N \cap L \neq \emptyset} m(N) \tag{29}$$

The plausibility and belief are related to each other as follows:

$$pl(L) = 1 - bel(\overline{L}) \quad (30)$$

Conversely for finite set L, we can denote the masses of m (L) with the following inverse function given the belief measure bel (N) for all subsets N of L.

$$m(L) = \sum_{N \mid N \subseteq L} (-1)^{|L-N|} bel(N) \quad (31)$$

where the difference of the cardinalities of the two sets is represented by |L − N|.

### 3.5.2. Dempster's rule of combination

The two sets of masses $m_1$ and $m_2$ are combined to depict joint mass and represented as below:

$$m_{1,2}(\emptyset) = 0 \quad (32)$$

$$m_{1,2}(L) = (m_1 \oplus m_2)(L) = \frac{1}{1-K} \sum_{N \cap C = A \neq \emptyset} m_1(N) m_2(C) \quad (33)$$

where

$$k = \sum_{N \cap C = \emptyset} m_1(N) m_2(C) \quad (34)$$

k is an estimate of the amount of conflict between the two mass sets.

The characteristics of DST has ignorance part such that probability of all events cumulative to 1. Ignorance is reduced in this theory by incorporating more and more evidences and combination rule. The rule is utilized to integrate various types of possibilities. This theory has much lower level of ignorance. Uncertainty interval can be reduced by adding more information. The disadvantage of this theory implies that if the computation is high, we have to deal with $2^n$ of sets.

### 3.5.3. Related Works based on DST

Fuzzy soft set-based decision making was improved by mitigating uncertainty especially in the medicine field by the authors of [72]. They utilized DST and ambiguity measures for this purpose. Their proposed approach proved to be efficient and feasible as it improved performance by reducing the uncertainty occurred by people's subjective cognition.

Authors in (Porebski et al. 2018) proposed a diagnosis support method and a rule selection both by applying fuzzy set and Dempster–Shafer theories. They utilized their model to diagnose liver fibrosis. They extracted information from a real dataset containing hepatitis C patients.

Authors of (Xiao 2018) proposed a hybrid framework utilizing DST with the belief entropy. The feasibility and efficacy of their method were validated by implementing a numerical example and a medical application.

Straszecka et al. (Straszecka 2006) introduced a unified fuzzy-probabilistic approach for medical diagnosis modelling processes. The basic concepts of DST, i.e. a basic probability assignment and focal elements correspond to the impact of an individual symptom in the diagnosis and disease symptoms, respectively. Focal elements' interpretation as fuzzy sets, evidence uncertainty, and imprecision of diagnosis were the novel approaches of their method.

Biswas et al. (Biswas et al. 2020) presented a novel decision-making strategy based on DSTby applying soft fuzzy sets for elucidation of pneumonia malformation in low-dose x-ray images. Gray level ambiguity was classified utilizing the S-function to define fuzzy soft sets and uncertainties were discriminated via fuzzy interval and Dempster-Shafer approach adapted for a decision criterion.

In (Ghasemi et al. 2013), authors combined Dempster–Shafer Theory and fuzzy inference system for brain MRI segmentation where the spatial information and the pixel intensities were utilized as features. The novelty of their work lies in the aspect that rules were paraphrased as evidences. The experimental results demonstrated that the proposed method called fuzzy Depster-Shafer inference system (FDSIS) exhibited competitive output using both real and simulated MRI databases.

The study in (Shi et al. 2018) integrated local classification model (LCM) to predict the drug-drug interactions via DST. Their supervised fusion rule combined the results from multiple LCMs. Their LCM-DS model exhibited better performance as compared to three different prevailing approaches.

Authors in (Kang et al. 2018) proposed a prognostic model based on DST and Gaussian mixture model for Clostridium difficile contagion. Criteria ratings of risk factors generated by the model helped the hospital administrators and risk managers to control and predict of Clostridium difficile infection prevalence.

Li et al. (Li et al. 2015) presented a framework based on fuzzy soft set and DST to help the clinicians in medical diagnosis. It was proved to be effective and feasible application in medical diagnosis.

Researcher in (Bloch 1996) exploited crucial features of DST in medical imaging. He pointed out the key aspects of the theory. It included the introduction of global or partial ignorance, the computation of conflict between images, and modelization of both imprecision and uncertainty in medical image processing. Partial volume effect in MR images can be managed properly by this approach.

Wang et al. (Wang et al. 2015) extended fuzzy DST to model domain knowledge using fuzzy and probabilistic uncertainty for medical diagnosis. A novel evidential structure to mitigate information loss was proposed. They presented novel intuitionistic fuzzy evidential reasoning (IFER) methodology that integrated inclusion measure and intuitionistic trapezoidal fuzzy numbers to enhance the accuracy of reasoning and representation.

Raza et al. (Raza et al. 2006) utilized DST to fuse the results of classification of breast cancer data from two sources: Fine-Needle Aspirate Cytology and gene-expression patterns in peripheral blood cells data. The classifiers used were support vector machine with polynomial, linear, and Radial Base Function kernels. The output of the fused classifiers yielded better results in detecting the breast cancer automatically. A brief summary of these works are summarized in Table 5.

*Table 5. Brief description of papers which used DST to handle uncertainty.*

| Authors | Year | disease/Area | Work description |
|---|---|---|---|
| Biswas et al. (Biswas et al. 2020) | 2020 | Pneumonia | Pneumonia malformation based on fuzzy soft set and DST |
| Xu et al. (Xu et al. 2020) | 2020 | Emotion recognition | Emotion recognition using DST |
| Prameswari et al. (Prameswari et al.) | 2019 | Digestive System Disorders | E-diagnostic model for the disorder of the digestive system using DST |
| Lima et al. (Lima and Islam 2019) | 2019 | Brain MRI segmentation | Brain MRI segmentation using a modified method |
| Razi et al. (Razi et al. 2019) | 2019 | Brain-computer interface (BCI) | multi-class motor imagery job classification of BCI |
| Soroush et al. (Zangeneh Soroush et al. 2019) | 2019 | Emotion recognition | Emotion recognition via DST and EEG |
| Laha et al. (Laha et al. 2019) | 2019 | EEG-induced Probabilistic Prediction | Color-Pathways in the Brain using DST |
| Porebski et al. (Porebski et al. 2018) | 2018 | Liver fibrosis | Automated liver fibrosis diagnosis |
| Xiao (Xiao 2018) | 2018 | Medical Diagnosis | A decision making algorithm in medical analytics using fuzzy soft sets and DST. |
| Shi et al. (Shi et al. 2018) | 2018 | Drug-drug interactions | Prediction of drug-drug interactions |
| Kang et al. (Kang et al. 2018) | 2018 | Clostridium difficile infection | A prognostic model for Clostridium difficile disease |
| Wang et al. (Wang et al. 2016) | 2016 | Medical Diagnosis | Ambiguity measure and DST in medical diagnosis |
| Li et al. (Li et al. 2015) | 2015 | Medical Diagnosis | DST and grey relational analysis based medical analytics |
| Wang et al. (Wang et al. 2015) | 2015 | Stroke | Intuitionistic fuzzy evidential reasoning (IFER) |

| Ghasemi et al. (Ghasemi et al. 2013) | 2013 | Brain MRI segmentation | Fuzzy DS inference model for brain MRI segmentation |
|---|---|---|---|
| Straszecka et al. (Straszecka 2006) | 2006 | Medical Diagnosis | The study of imprecision and uncertainty in models of medical diagnosis |
| Raza et al. (Raza et al. 2006) | 2006 | breast cancer | Prediction of breast cancer tumors |
| Bloch (Bloch 1996) | 1996 | Medical Imaging | Multi-modality medical images classification |

## 3.6. Imprecise probability

Imprecise probability can be achieved by generalizing traditional probability. A set of probabilities with lower and upper probabilities Prob($Z$)=[$p_1$,$p_2$] are applied instead of using probabilistic measure Prob($Z$)=$p$ related with an event Z for quantification both epistemic and aleatory uncertainties [84]. Several representations and theories of imprecise probability have been devised such as the coherent lower prevision theory, possibility theory, de Finetti's subjective probability theory, and Dempster-Shafer evidence theory.

### 3.6.1. Definitions

Consider upper probability $\overline{P}$ (Z) and lower probability $\underline{P}$ (Z), with $0 \leq \underline{P}(Z) \leq \overline{P}(Z) \leq 1$. There is precise probability if $\overline{P}$ (Z) = $\underline{P}$ (Z) = P(Z) for all events A. There is lack of knowledge completely about Z if $\overline{P}$ (Z) =1 and $\underline{P}$ (Z) = 0. For disjoint events Z and Y:

$$\underline{P}(Z \cup Y) \geq \underline{P}(Z) + \underline{P}(Y) \text{ and } \overline{P}(Z \cup Y) \leq \overline{P}(Z) + P(Y) \tag{35}$$

$$\underline{P}(not - Z) = 1 - \overline{P}(Z) \tag{36}$$

Precise probability distributions of closed convex set $\mathcal{P}$:

$$\underline{P}(Z) = inf_{p \in \mathcal{P}} p(Z) \text{ and } \overline{P}(Z) = sup_{p \in \mathcal{P}} p(Z) \tag{37}$$

Subjective interpretation:

$\overline{P}$ (Z): minimum cost at which selling the gamble is desirable

$\underline{P}$ (Z): maximum cost at which buying gamble paying 1 if Z occurs and else 0 is desirable

$\underline{P}$ (Z) can be deduced as reflecting the evidence in favor of event Z, 1- $\overline{P}$ (Z) as reflecting the evidence against Z, in favor of not-Z.

Imprecision Δ (Z) = $\overline{P}$ (Z) - $\underline{P}$ (Z) reflects lack of perfect information about probability of Z.

### 3.6.2. Related Works on Imprecise Probability

Coletti et al. (Coletti and Scozzafava 2000) presented the role of coherence in handling and eliciting imprecise probabilities and its application to medical analytics. They also focused on the distinction between syntactic and semantic aspects. Mahmoud et al. (Mahmoud 2016) used different machine learning decision tree classification algorithms on noisy medical datasets. They employed three decision tree methods: single tree classifiers, ensemble models, and creedal decision trees (CDTs) to tackle the uncertainty measures and imprecise probabilities. CDTs outperformed other two methods in noisy environments. Van Wyk et al. (Van Wyk 2020) investigated the medicinal plants in African region. They studied traditional diversity patterns in selection of medicinal plants of Africa. They utilized imprecise Dirichlet model with linear regression and Bayesian analysis.

Kwiatkowska et al. (Kwiatkowska et al. 2009) revisited the concept of uncertainty, incompleteness, and imperfection of data. They studied the traditional hierarchical approach to knowledge, information, and data with reference to medical data, which is characterized by time-dependency, variable granularity and heterogeneity. They argued that it was contextual to interpret the imprecision and medical data could not be decoupled from their intended usage. They integrated multidimensional, fuzzy-logic, and semiotic approaches to propose a framework for medical data modelling to address contextual interpretation of imprecision issues. A brief description of these papers are summarized in Table 6.

*Table 6. Brief description of papers which used imprecise probability to handle uncertainty.*

| Authors | Year | Disease/Area | Work description |
|---|---|---|---|
| Van Wyk et al. (Van Wyk 2020) | 2020 | Traditional African Medicine | Study of medicinal plants used in Traditional African Medicine |
| Jafar et al. (Jafar et al. 2020) | 2020 | Medical Diagnosis | Neutrosophic soft sets in medical diagnosis |
| Meilia et al. (Meilia et al. 2020) | 2020 | forensic medicine | causal inference in forensic medicine |
| Kosheleva et al. (Kosheleva and Kreinovich 2019) | 2019 | Natural Next Approximations | Natural Next Approximations to imprecise probabilities |
| McKenna et al. (McKenna et al. 2018) | 2018 | Breast Cancer | Modelling for chemotherapy in Breast Cancer |
| Gandhimathi et al. (Gandhimathi 2018) | 2018 | Medical Diagnosis | Fuzzy soft matrix in medical diagnosis |
| Ayalon et al. (Levis et al. 2018) | 2018 | Depression Diagnosis | Classification of major depression diagnosis |
| Mahmoud et al. (Mahmoud 2016) | 2016 | Thrombosis, Heart Disease, Arrhythmia, Hypothyroid | Diagnosing noisy medical data by applying various Intelligent Tree Based Classifiers |
| Kwiatkowska et al. (Kwiatkowska et al. 2009) | 2009 | Sleepiness measures | Interpretation of Imprecision in Medical Data |
| Coletti et al. (Coletti and Scozzafava 2000) | 2000 | Infectious diseases | Coherence in handling imprecise probabilities and its utilization in medical diagnosis |

## 4. Discussion

Studying uncertainty involves uncertainty in data and uncertainty in the model. Data uncertainty arises from sources such as measurement noise, transmission noise, and missing values. Model uncertainty comprises of not knowing the best architecture and parameters which can predict future data. Uncertainty quantification helps to enhance the confidence in the results obtained by different methods. Nowadays, the growth of new technologies has paved the way to produce huge amounts of raw data in different fields. The use of such raw data is never easy as it might include noise. There are few researches like (Oh et al. 2018) that do not require noise filtering because of end to end training of deep learning network using noisy ECG signals. We know that CNN is less sensitive to noise. It can extract information even when the data are noisy (Qian et al. 2016). In another research (Karimifard and Ahmadian 2011), Hermitian basis functions were used to extract higher order cumulants of the ECG beats. They could reduce the effects of Gaussian noise. However, commonly before getting useful information from data, they need to be cleaned for the presence of any noise and unnecessary information. Hence, dealing with uncertainty in both data and model is an important subject for researchers to make better decisions in various domains. Thus, researchers across the world are trying to deal with uncertainty using machine learning algorithms and probability theories. We have discussed algorithms namely Bayesian inference, fuzzy systems, Monte Carlo simulation, rough classification, Dempster–Shafer theory, and imprecise probability to handle the uncertainties. In Figure 8, the percentage of uncertainty handling algorithms used in medical field by researches is shown.

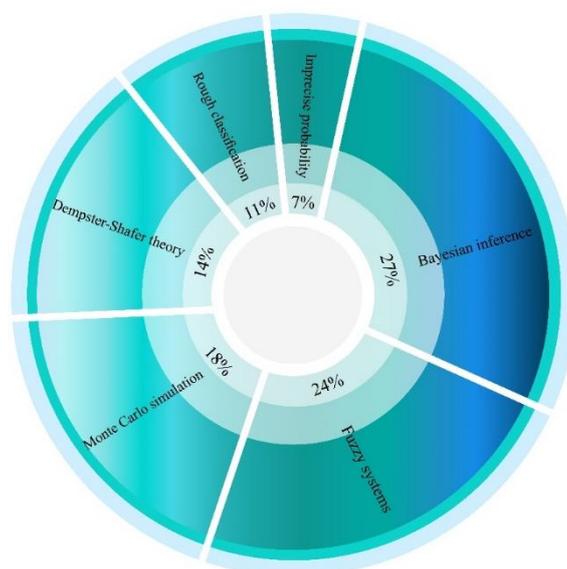

*Figure 8. Percentage of uncertainty handling algorithms used in medical field researches.*

It can be noted from Figure 8 that, Bayesian inference, fuzzy systems, and Monte Carlo simulation respectively have been used by many researchers in the medical field. Among these methods, Bayesian inference is more widely used as compared to other methods (Dempster 1968; Ma et al. 2006; Minka 2001). The involvement of average of parameters makes this method self-regularized. Both uncertainty types are tackled by this method. Prior knowledge is included while using Bayesian inference. But it suffers from few disadvantages such as it is computationally intensive. The intractability of its integrals and high dimension are other disadvantages of this method. Fuzzy systems are the second most widely used technique. These methods are simple and also can overcome different types of uncertainty efficiently (Kosko 1994). MCS has few advantages such as it can address the intractability analytically and can survey the parameter space of a problem completely. Meanwhile, the MCS results are relatively easy to understand, flexible and empirical distributions can be handled. The limitation of MCS is its computational cost and solutions are not exact. They depend on the number of repeated runs (Mooney 1997). Dempster–Shafer theory in combination with fuzzy systems also attracted the attention of many researchers (Wang et al. 2016). The main disadvantage of the rough set (Dash and Patra 2013) is that it cannot handle real-valued data. Using a real-valued rough set is the best alternative to solve this problem. Meanwhile, fuzzy-rough set theory can be used to handle this problem (Jensen and Shen 2004; Zhai 2011). A brief description of the pros and cons of the methods used are summarized in Table 7.

Few other methods can also be used to tackle the uncertainty quantification. Some of them are Bayesian information criterion (Pho et al. 2019), Laplace approximation (Friston et al. 2007), variational approximations (Ormerod and Wand 2010), exact sampling (Propp and Wilson 1996), and expectation propagation (Minka 2001). However, these methods are not much used by the researches.

There are also some challenges when investigating this topic. Some of them are: 1) handling dimensionality and computational cost of the problems, 2) testing the validity of different methods which can be used to handle uncertainty, 3) testing the validity of achieved results, 4) adaptation of model inputs to specific patient setting, and finally 5) determining the model complexity level which yielded more accurate outcomes.

In the fourth challenge, the inputs may include the computational domain, boundary conditions, and physical parameters. Measurement uncertainty and large biological variability lead to hampering of all measurements of these inputs. Consequently, it leads to uncertainties in the inputs. Meanwhile, some of the model inputs are unmeasurable or they are measurable but costly. The model complexity increases if it has to describe the reality more accurately is the fifth challenge. Consequently, it lessens the uncertainty in the output of the model. However, this enhances the uncertainty in the input of the system simultaneously because more inputs of the system ought to be examined patient-specifically. As illustrated in Figure 9, it is necessary to find an optimal transaction between the uncertainty resulting from system and data that can initiate nominal total uncertainty (Huberts et al. 2014). To deal with challenges number four and five, it is essential to measure the output uncertainty of the system. The uncertainties in the inputs of the system are responsible for it. The pros and cons of it developed to handle the uncertainties are summarized in Table 7.

*Table 7. Advantages and disadvantages of the approach devised to handle the uncertainities.*

| Method | Advantages | Disadvantages |
| --- | --- | --- |
| ANFIS (Pandya 2015) | <ul><li>Enhances the fuzzy if- then rules without human expertise to delineate the behaviour of a complex system</li><li>Fast convergence time</li><li>Utilizes membership functions</li><li>Rapid learning capacity</li><li>Ability to capture non-linear structure of a process</li></ul> | <ul><li>Spatial exponential complexity</li><li>Coefficients signs not all the time steady with basic monotonic relations</li><li>Symmetric fault handling and high outliers control</li><li>Limitations of the number of inputs</li></ul> |
| Bayesian inference | <ul><li>Involvement of average parameters make the method as self-regularized</li><li>Including good information should enhance prediction</li><li>It takes care of both types of uncertainty</li><li>Prior knowledge is effortlessly integrated.</li></ul> | <ul><li>Computationally challenging.</li><li>Integrals are inflexible and elevated dimension.</li><li>There can be too many hypotheses</li><li>If the prior information is wrong, it can send inferences in the wrong direction.</li></ul> |
| Rough set | <ul><li>Does not need any additional information or preliminary about data like probability distributions which is needed in statistics</li><li>Generates a set of decision rules from data</li><li>Easy-to-understand formulation.</li></ul> | <ul><li>Cannot handle real-valued data.</li></ul> |
| MCS | <ul><li>Can address questions that are intractable analytically.</li><li>Surveys the parameter space of a problem completely.</li><li>Results are relatively easy to understand.</li><li>Empirical distributions can be handled.</li></ul> | <ul><li>High computational cost.</li><li>Solutions are not exact.</li></ul> |
| **DST (Parikh et al. 2001)** | • It has a lower level of ignorance<br>• Uncertainty interval reduces as more intervals are added<br>• Multi-criteria decision making can be achieved by combining DST with Analytic Hierarchy Process<br>• DST as a theory of evidence has to account for the combination of different sources of evidence | • Computation is high as we need to deal with 2n sets. |
| **Imprecise Probability (Coolen et al.** | • Possible to tackle conflicting evidence<br>• Useful in critical decision problems<br>• Imprecise probability theory gives a generalization of possibility theory and evidence theory. It also lets us to understand some results of these approaches in reputable analysis. | • There is commonly an independent degree of caution or boldness inherent using one interval, rather than a wider or narrower one. This may be a degree of confidence, threshold of acceptance, or degree of fuzzy membership.<br>• Computational burden is usually higher |

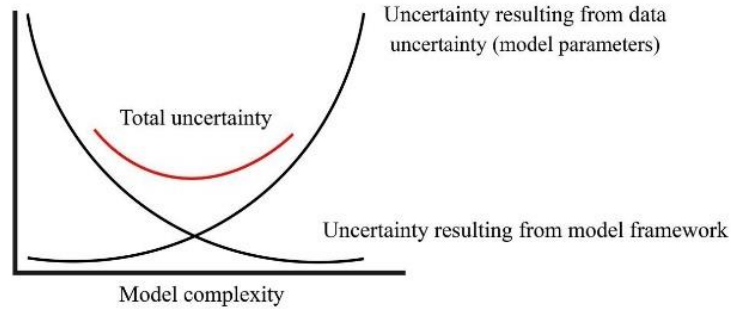

*Figure 9. The trade-off between data and model uncertainty.*

Higher order statistics/spectra (HOS) (Chua et al. 2010) and deep learning (DL) techniques are found to more robust to noise. The bispectrum and cumulants widely used for medical data to capture the subtle changes in the signal (Pham et al. 2020). It can detect deviations from linearity, stationarity or Gaussianity in the signal. In nature, the biomedical signals are commonly non-stationary, non-linear and non-Gaussian. So, it is better to analyse them with HOS rather than the use of second-order correlations.

We know that data collection without any noise in the data is expensive and time-consuming. Nowadays, due to the progress in storage devices and technologies, a huge amount of medical data are available. These medical data are noisy and need to be eliminated to get an accurate diagnosis. HOS and deep learning techniques are found to be suitable to handle this type of noisy data. Deep neural networks can generalize on noisy data in training, instead of just memorizing them (Rolnick et al. 2017). In addition, it is shown in (Rolnick et al. 2017), deep neural networks can learn from data that has been impacted by an arbitrary amount of noise. By applying deep neural networks on multiple datasets such as MINST, CIFAR-10 and ImageNet, the authors showed that successful learning is possible even with an essentially arbitrary amount of noise. Authors have showed that using deep learning techniques, it is possible to get excellent performance even with noisy electrocardiogram (ECG) signals (Acharya et al. 2017; Gao et al. 2019; Jeong et al. 2019; Oh et al. 2018; Oh et al. 2019). Hence, in future, DL models may be explored to nullify the presence of noise in the medical data.

## 5. Conclusions and Future works

This paper reviewed uncertainty quantification challenges in the medicine/health domain. Unlike industrial fields, the medical field has always been determined by several factors leading to uncertainty in decisions and outcomes.

More useful methods of handling uncertainty in healthcare should start with physicians. To achieve this goal, first, doctors need to have a clear and better picture of uncertainty in their work and strive for novel approaches to handle this challenge. Decision making in this field is intricate. So, it needs intuitive and rational thinking. Although uncertainty is an inevitable factor in medical decisions, clinicians commonly downplay its importance. Unfortunately, it is not a suitable response to this type of challenges. Most of the issues about conventional outlooks and practices stem from persons who have not earlier been engaged in the decision-making of medical services. Unfortunately, the ways of the respond from the medical professional to the uncertainty during their job may lead to damaging upshots on complex decision-making processes.

We know that artificial intelligence methods automate processes and make them available more widely. However, it is not being used to its full potential in the medical field. We need uncertainty quantification because it leads to early detection of diseases, save time, and accurate diagnosis.

The findings of this review paper demonstrated that different types of classical machine learning and probability theory techniques have been significantly used in uncertainty classification. Recently deep learning techniques are becoming very popular among researchers due to its high performance. Using deep learning methods even with noisy signals, high arrhythmia classification was obtained (Acharya et al. 2017; Oh et al. 2018; Oh et al. 2019). So, in future, such techniques can be employed to handle the uncertainty in the medical data and obtain high performance.

## Data availability

The authors declare that all data supporting the findings of this study are available within the paper and its Supplementary Information.

## Declaration of interests

We declare no competing interests.

## Funding

The authors have not declared a specific grant for this research from any funding agency.